# DESIGN FOR A BLIND STEREOSCOPIC PICTURE TAKER

D. MOCTEZUMA-ENRIQUEZ[1], E. RODARTE-LEYVA.

*[1] Intstituto Tecnológico de Estudios Superiores de Monterrey Campus Sonora Norte.*
*Blvd. Enrique Mazón López 965, 83000 Hermosillo, Son.*
*E-mail: dmoctezu@tec.mx*
*Rosales y Blvd. Luis Encinas, Hermosillo, Sonora 83190, Mexico.*

Abstract. An Schematical Design for an Autonomous Picture taker used for obtaining Point clouds from pictures taken inside a House. In this case we are proposing the use of an equation programmed inside an embedded system that will be tracking the points inside a room and then, open the space between two cameras of same type in order to take pictures that later will be used to create the cloud points for the mathematical model that the latter user will apply to that pictures.

## 1. INTRODUCTION

Two-dimensional images of a three-dimensional body photographed by two video cameras are parallactically compensated prior performing the detection of the corresponding point necessary to recognize the body. Namely, the parallax between both cameras for the object distance measured by a range finder is calculated by a parallax calculator, and one image is electrically shifted by an image shifting circuit by only the number of pixels in accordance with the parallax value, thereby performing the parallax compensation of the images. In a calculation processing unit, the correlation values for the corresponding points between both images are calculated, and the distance images of the body are produced on the basis of these correlation values.

This was the acceptable example of what is called an Stereoscopic device, in this case the multiple usages on stereoscopic systems recalls, the virtual reality and the image analysis used on gravimetry and image processing.

Although this technique is quite interesting, for obtaining measurements and patterns, modifications to the image capture systems may allow reducing the number of variables with which to build the mathematical model and therefore the computing power can be rationalized. In this case it would be the use of a stereoscopic camera system.

The stereoscopic 3D images include a pair of images taken with a binocular device. Humans use the disparity in such stereoscopic images to determine depth, through the principle of stereopsis. The two images in a stereoscopic pair are not arbitrary images, but contain a consistent image content related to the disparity, which depends on the depth [1]. The editing of the images independently using the best human efforts is likely to break the stereoscopic consistency; even minor discrepancies lead to a bad stereoscopic viewing experience or deceptive depth information. For example, an anaglyph image can not rotate directly, since this would introduce a vertical disparity, giving an unnatural stereoscopic experience [1].

Given two images captured by a binocular device, a direct approach to the manipulation of the stereoscopic perspective would be to calculate the depth in each pixel from the pair of stereo images, reconstruct the geometry of the scene and then calculate the projection in the new configuration of the camera. However, this solution would not be perfect, since the occlusion would generate gaps in the output. In addition, while the human visual system can easily understand 3D geometry from a pair of stereo images, it is much more difficult to algorithmically reconstruct the geometry of the scene accurately, even with the help of user interaction, and is needed a more robust solution.

## 2 THEORY

The principles of perspective representation, where aview of threedimensional (3-D) space is mapped onto a 2-D picture window, has been attributed to Alberti in the fifteenth century [2]. This simple procedure is an additional source of spatial error when translated to stereoscopic displays. The correct presentation of disparity cues is critically dependent on the half-image separation, image plane distance and the user's interpupillary distance (IPD). In traditional desk mounted stereoscopes these factors are assumed to be stable and the principles for correct half-image placement have been well documented. Vergence eye-movements, however, introduce small changes in the inter-ocular distance (IOD) because the centre of ocular rotation lies some 5-6 mm behind the nodal point of the optical system [3]

As can be seen in the cases of interest, like th usage of 3D glasses or the Virtual reality, even on magazine games where trhough overlapping images we can build an third image, the results of the use of stereoscopy in the treatment of images can be observed. And as the changes in the separation of the lenses can cause interesting effects in pictures. From the positions of the cameras the separation and the angles of comparison the existing anaglyph (the image distorted in colors) the information obtained subjected to a mathematical treatment will result in the measurement of depth applied to know the proximity or distance of objects inside of the three-dimensional model.

From here the essential problem is then observed, what is the optimal parameter?

The panels of the right and left eye in a stereoscopic reconstruction are created by projection from the main points of the twin recording chamber.

The geometrical situation is more clearly understood when analyzing how screens are generated when photographing a small cubic element of lateral length $dx = dy = dz$ from a distance z with a twin camera whose lenses are separated by a distance. As seen in Fig. 1.

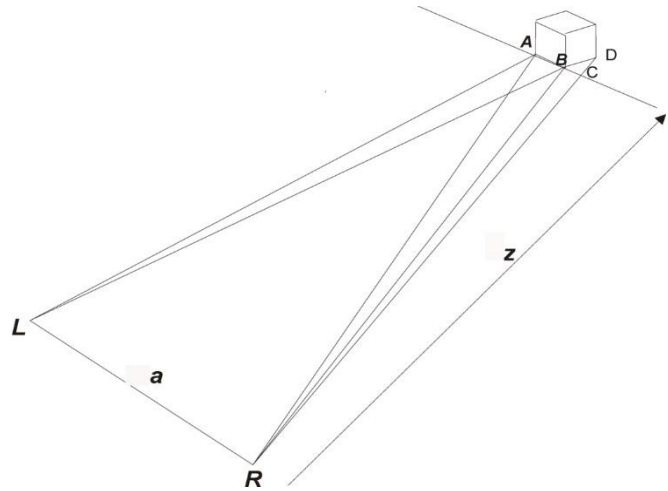

Fig. 1 Adjustments to be made for separation glasses in cameras and build an stereoscopical device.

In the left eye panel of the stereogram, the distance *AB* is the representation of the front face of the cube, in the light eye panel, there is also BC, the representation of the depth of the cube, that is, the intersection in the screen of the rays from the main points of the cameras towards the back of the cube. This interval is calculated in the first order for *dz (a / z)*. (To simplify the account, the right and left screens overlap, as they would be on a 3D screen with glasses

LCD). Hence the depth / width relationship of the cube view, as represented in its representation on the display screen, is $r = (a \times dz) / (z \times dx) = a / z$ since $dx = dz$ and it depends only on the distance of the objective of the twin lenses and their separation and remains constant with changes of scale or magnification. The relation *(depth / width)* of the real object, of course, is 1.00.

## 3 BUILDING AND OPERATION

In this case a three-dimensional design was made for the construction of this device, the mechanical model used as a base a gear system that allowed the independent movement of the base of 360 degrees rotation, as well as the use of a base of autonomous separation of the cameras this with the purpose of making a shot of hundreds of frames the system has a distance sensor so that according to the formula mentioned in the previous section the system take the pictures as well as self-adjust to points where the depth its going to change. This can be seen on design of Fig. 2.

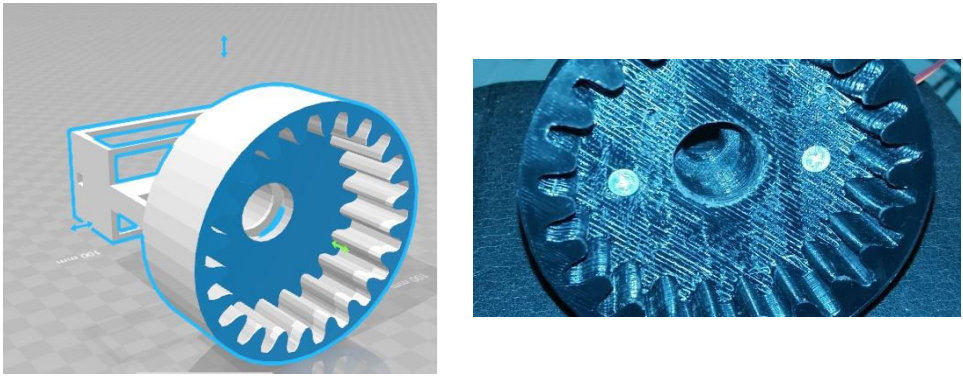

Fig.2 System designed, and system builded

On Fig.3 we can see an picture of the builded apparatus, and the picture of the builded separation base.

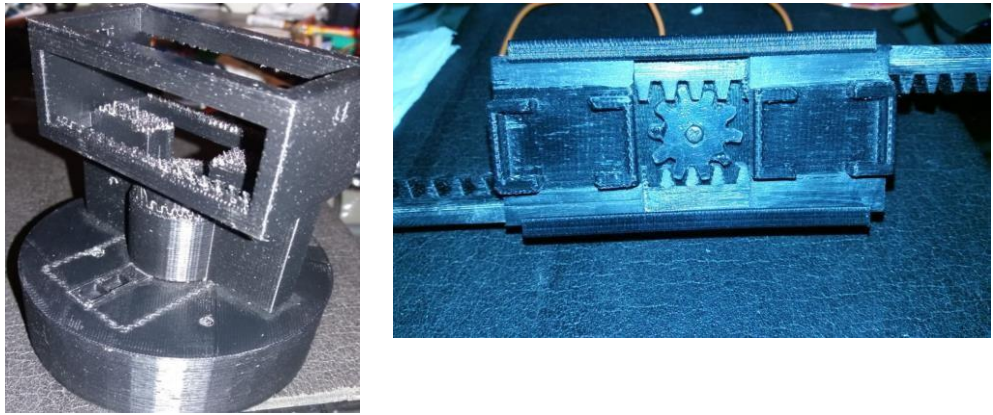

Fig.3. Builded 3D design.        And separation base.

The capture system works in essence in the following way:

1. The microcontroller system will launch an output signal for the distance transmitter, hence the reception signal serivá to launch pulses that will give the opening or lock depending on the distance to the opening reference value. As it is observed in the scheme of Fig. 5.

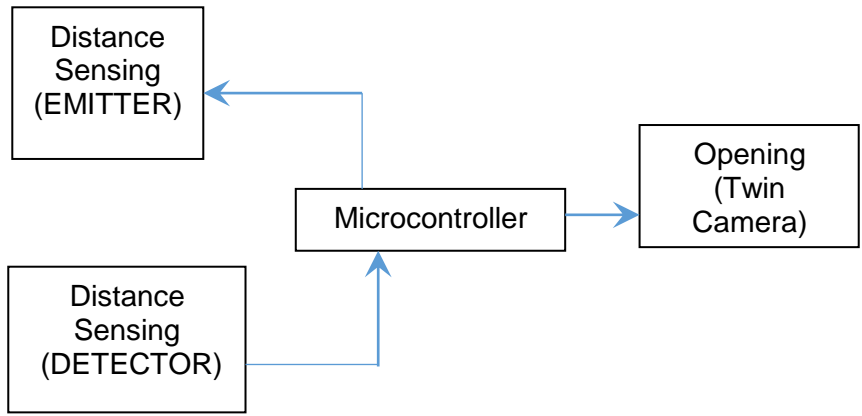

Fig.6. Basic Circuit Function.

The microcontroller system will evaluate the ignition moment as an advance marker. From there, the microcontroller will send a quantity of PWM pulses in a deretrminated direction so that the base moves a certain amount of degrees, at the moment when this completes a complete turn the system will finish the task, as seen in the diagram of Fig.

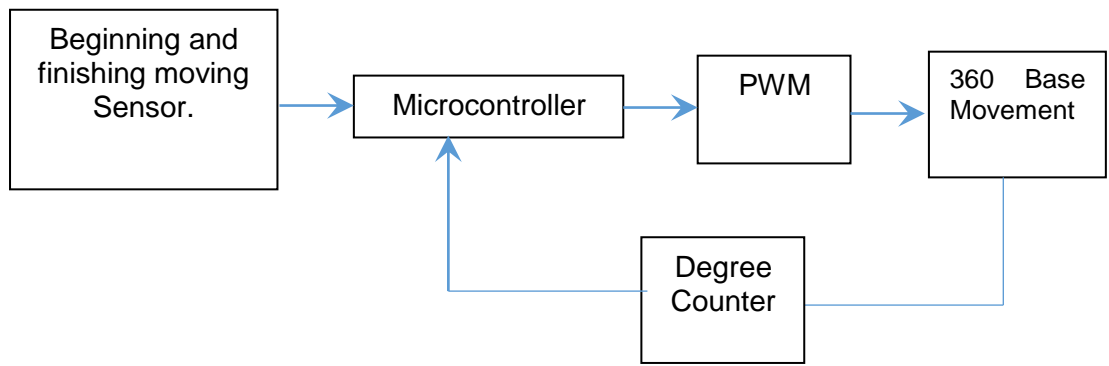

Fig.7. Base movement.

## 3 TEST

A preliminary construction was made using as reference the design with the considerations expressed in this document, of this construction the following tests were carried out.

1. Basic operation test:

In this test the value of the data taken by the distance sensor was taken into account and the opening values were checked in order to find the pulse rate and frequency for the PWM configuration.

2. Positioning test for overlap.

One of the interesting problems for the construction of panoramas and point clouds is the overlay aspect of photographs. For this case what should be done is to consider the opening and a rotation pattern, so for automatic control, it is necessary to perform a PWM in which both frequency and pulse width must be calibrated to obtain the best movement .

This test in the first instance must be made to trial and error in order to find an appropriate value of movement. This is for the purpose of making the movement automatically.

The important thing to find in these tests are values that will be in the microcontroller that will allow them to make calculations for the algorithms and that will end up giving a frequency and width value.

## 4 TEST RESULTS

With regard to test 1, it was obtained that a pulse of around 1333Hz with a 33% ontime allows to perform the movement of the opening of the camera with a movement rate of around 5mm / pulse.

Regarding test 2, it was obtained that there is a movement of at least about 5 ° per pulse at the moment of moving the motor, which implies that changes must be made in the orbital gear teeth in order to improve the photographic taking and make changes in the overlap. Before adding the cameras, On the mount.

# 5 CONCLUSIONS

In the advances of this stage, it was possible to specify the construction of a first prototype of tests, and the following observations were obtained:

1.      In the calculations made for the design, the 3D printing made with an XYZ printing machine DaVinci model 1.0 presents error margins of at least 3% at the time of taking measurements and comparing them with those designed on the platform.

2.      The measurements of the number of teeth with respect to the movement and pulses of PWM must be recalibrated in order to have a better value of shots and increase the number of them for the constitution of the used panorama with which the cloud of points will be made and it will be placed in the mathematical algorithm.

The work to be done in the next few months that follows for the constitution of a complete prototype will be:

1.      Readjustment of the angular measurements for the orbital gear and the change of the gears that are connected to the motor.

2.      Calibration of the modulated pulses in order that the new teeth adjust to the system and improve the photographic taking.

# 6 REFERENCES

[1] Bajura, M., Fuchs, H. & Ohbuchi, R. (1992). Merging virtual objects with the real world: Seeing ultrasound imagery within the patient. Computer Graphics, 26, 203-210.

[2] Cutting, J. E. (1986). Perception with an eye for motion, Cambridge, Mass: MIT Press. Fisher, S. K. & Ciuffreda, K. J. (1990). Adaptation to opticallyincreased interoeular separation under naturalistic viewing conditions. Perception, 19, 171-180.

[3] Bennett, A. G. & .Francis, J. L. (1962). In Davson, H. (Ed.), The eye (Vol. 4), New York: Academic Press.

[4] Cruz-Neira, C., Sandin, D. J. & Defanti, T. A. (1993). Surround-screen projection-based virtual reality: The design and implementation of the CAVE. Computer Graphics, 27, 135-142.

[5] Henson, D. B. & Dharamshi, B. G. (1982). Oculomotor adaptation to induced heterophoria and anisometropia. Investigative Ophthalmology and Visual Science, 22, 234-240. Judge, S. J. & Miles, F. A. (1985).

[6] Changes in the coupling between accommodation and vergence eye movements (the stimulus AC/A ratio) induced in human subjects by altering the effective interocular separation. Perception, 14, 617-629.

[7] Lemij, H. G. & Collewijn, H. (1991a). Long-term non-conjugate adaptation of human saccades to anisometropic spectacles. Vision Research, 31, 1939-1954.

[8] Maxwell, J. S. & Schor, C. M. (1994). Mechanisms of vertical phoria adaptation revealed by time-course and two-dimensional spatiotopic maps. Vision Research, 34, 241-251.

[9] Mon-Williams, M., Wann, J. & Rushton, S. (1993). Binocular vision in a virtual world: Visual deficits following the wearing of a head-mounted display. Ophthalmic and Physiological Optics, 13, 387-391.